\documentclass{article}
\usepackage{graphicx,amsmath,amsfonts,tikz,bm,nips12submit_e,times,xcolor}
\usepackage[pdftex]{hyperref}

\usetikzlibrary{arrows,positioning} 

\tikzset{
  select/.style={draw=black,line width=1.2mm},
  punkt/.style={circle,draw=black,minimum height=4em,text centered}
}

\newcommand{\f}[1]{\textbf{#1}}

\begin{document}

\title{Semi-Supervised Domain Adaptation with Non-Parametric Copulas}

\author{
David Lopez-Paz\\
MPI for Intelligent Systems\\
\vspace{0.5cm}
\texttt{dlopez@tue.mpg.de}\\
\And
Jos\'e Miguel Hern\'andez-Lobato\\
University of Cambridge\\
\texttt{jmh233@cam.ac.uk}\\
\And
Bernhard Sch\"olkopf\\
MPI for Intelligent Systems\\
\texttt{bs@tue.mpg.de}
}

\nipsfinalcopy

\maketitle

\begin{abstract}
A new framework based on the theory of copulas is proposed to address
semi-supervised domain adaptation problems.  The presented method factorizes
any multivariate density into a product of marginal distributions and bivariate
copula functions. Therefore, changes in each of these factors can be detected
and corrected to adapt a density model accross different learning domains.
Importantly, we introduce a novel vine copula model, which allows for this
factorization in a non-parametric manner.  Experimental results on regression
problems with real-world data illustrate the efficacy of the proposed approach
when compared to state-of-the-art techniques.
\end{abstract}

\section{Introduction}

When humans address a new learning problem, they often use knowledge acquired
while learning different but related tasks in the past.  For example, when
  learning a second language, people rely on grammar rules and word derivations
  from their mother tongue.  This is called \emph{language transfer}
  \cite{language_transfer}. However, in machine learning, most of the
  traditional methods are not able to exploit similarities between different
  learning tasks. These techniques only achieve good performance when the data
  distribution is stable between training and test phases. When this is not the
  case, it is necessary to a) collect and label additional data and b) re-run
  the learning algorithm.  However, these operations are not affordable in most
  practical scenarios.

\emph{Domain adaptation}, \emph{transfer learning} or \emph{multitask learning}
frameworks \cite{Mansour99,shai,Cortes11,survey} confront these issues by first, building a notion of
\emph{task relatedness} and second, providing mechanisms to \emph{transfer
knowledge} between similar tasks.  Generally, we are interested in improving
predictive performance on a \emph{target task} by using knowledge obtained when
solving another related \emph{source task}. Domain adaptation methods are
concerned about \emph{what} knowledge we can share between different tasks,
\emph{how} we can transfer this knowledge and \emph{when} we should do it or
not to avoid additional damage \cite{bin}.

In this work, we study semi-supervised domain adaptation for regression tasks.
In these problems, the object of interest (the mechanism that maps a set of
inputs to a set of outputs) can be stated as a conditional density function.
The data available for solving each learning task is assumed to be sampled from
modified versions of a common multivariate distribution. Therefore,  we are
interested in sharing the ``common pieces'' of this generative model between
tasks, and use the data from each individual task to detect, learn and adapt
the varying parts of the model. To do so, we must find a decomposition of
multivariate distributions into simpler building blocks that may be studied
separately across different domains.  The theory of copulas provides such
representations \cite{nelsen}.

Copulas are statistical tools that factorize multivariate distributions into
the product of its marginals and a function that captures any possible form of
dependence among them. This function is referred to as the copula, and it links
the marginals together into the joint multivariate model. Firstly introduced by
Sklar \cite{sklar}, copulas have been successfully used in a wide range of
applications, including finance, time series or natural phenomena modeling
\cite{copula_applications}.  Recently, a new family of copulas named
\emph{vines} have gained interest in the statistics literature \cite{vines}.
These are methods that factorize multivariate densities into a product of
marginal distributions and bivariate copula functions.  Each of these factors
corresponds to one of the building blocks that we assume either constant or
varying across different learning domains.

The contributions of this paper are two-fold. First, we propose a
non-parametric vine copula model which can be used as a high-dimensional
density estimator. Second, by making use of this method, we present a new
framework to address semi-supervised domain adaptation problems, which 
performance is validated in a series of experiments with real-world data and
competing state-of-the-art techniques.

The rest of the paper is organized as follows: Section \ref{sec:copulas}
provides a brief introduction to copulas, and describes a non-parametric
estimator for the bivariate case.  Section \ref{sec:vines} introduces a novel
non-parametric vine copula model, which is formed by the
described bivariate non-parametric copulas.  Section
\ref{sec:domain_adaptation} describes a new framework to address
semi-supervised domain adaptation problems using the proposed vine method.
Finally, section \ref{sec:experiments} describes a series of experiments that
validate the proposed approach on regression problems with real-world data.

\section{Copulas}\label{sec:copulas}
\begin{figure}
  \begin{center}
      {\includegraphics[width=0.28\linewidth]{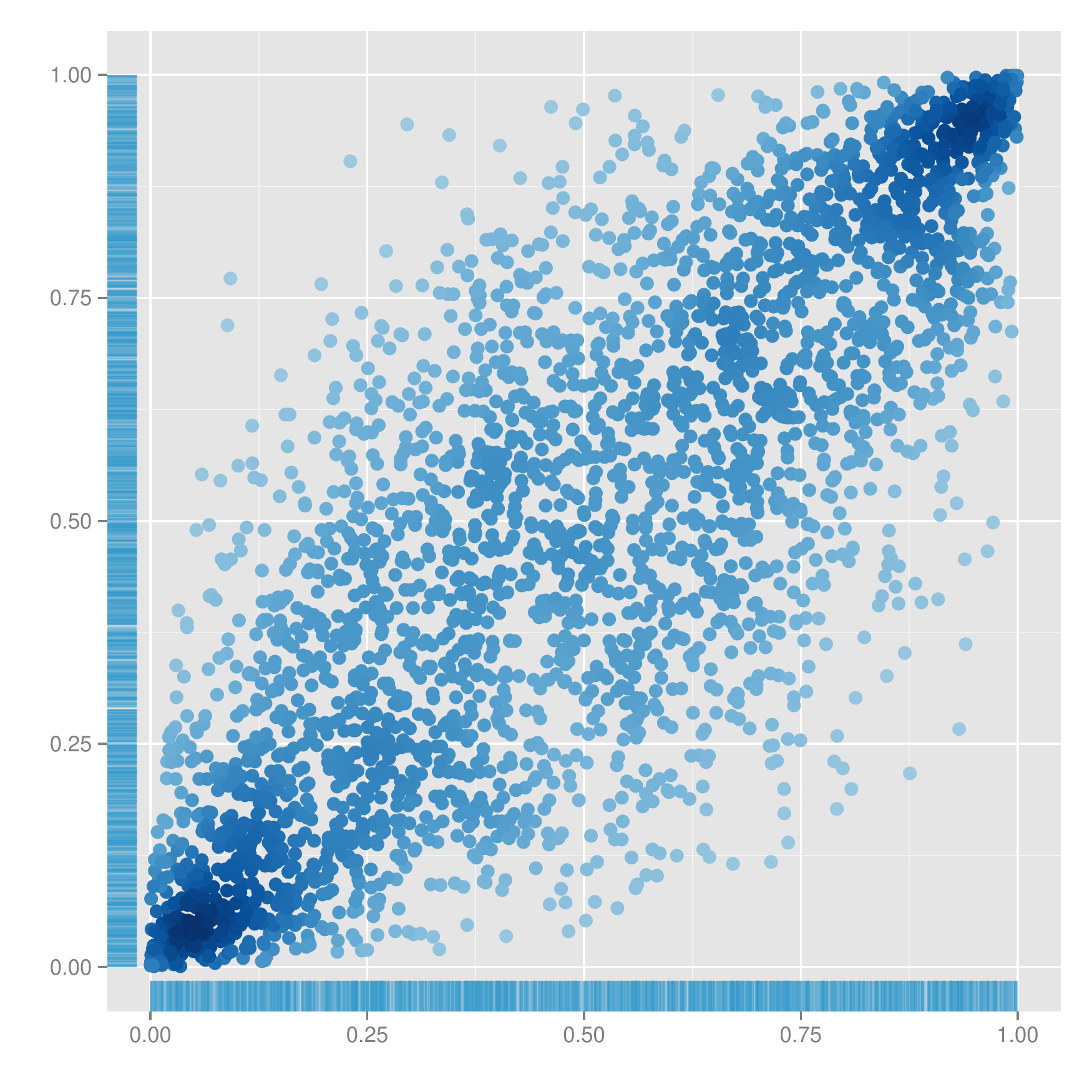}}\hskip 1 cm
      {\includegraphics[width=0.28\linewidth]{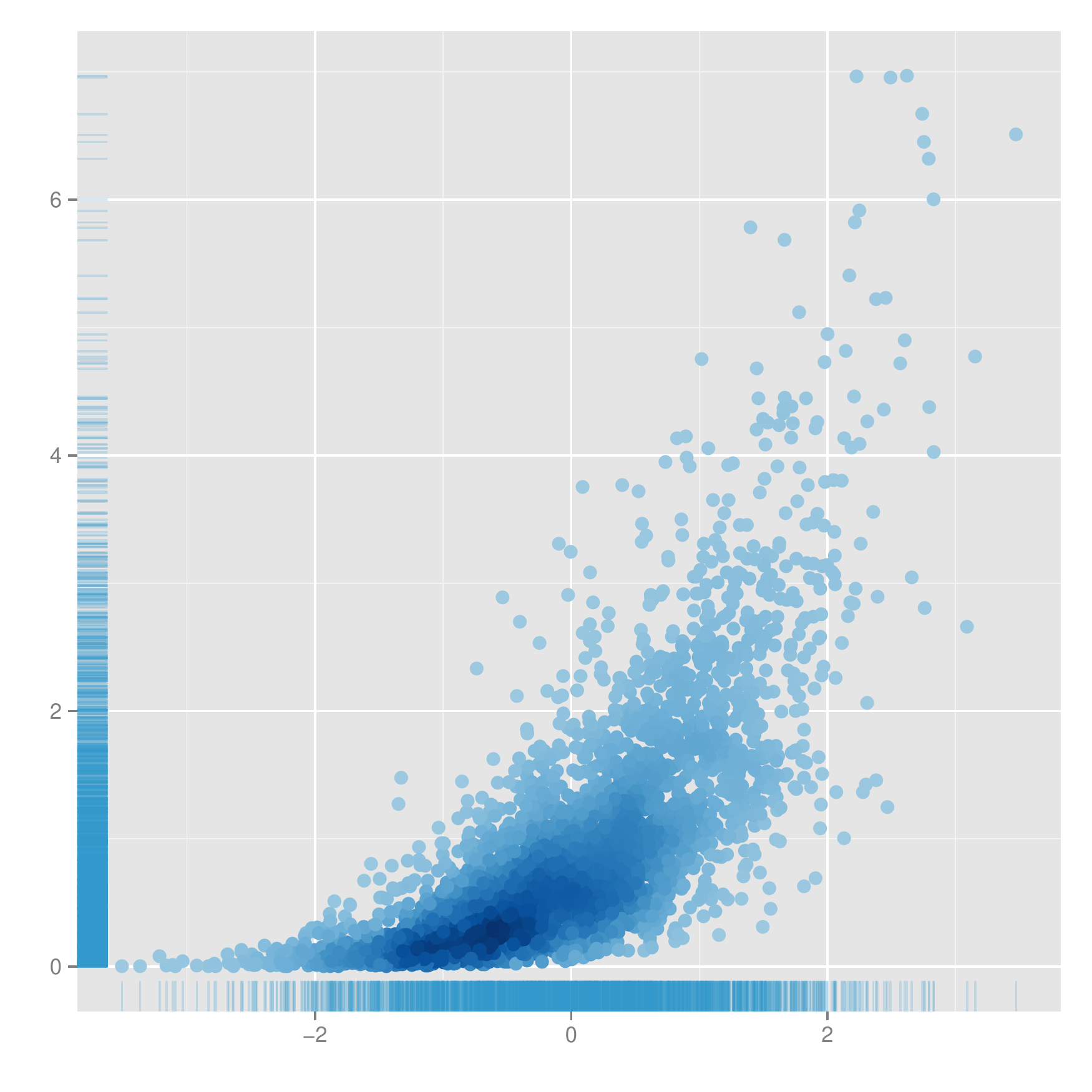}}\hskip 1 cm
      {\includegraphics[width=0.28\linewidth]{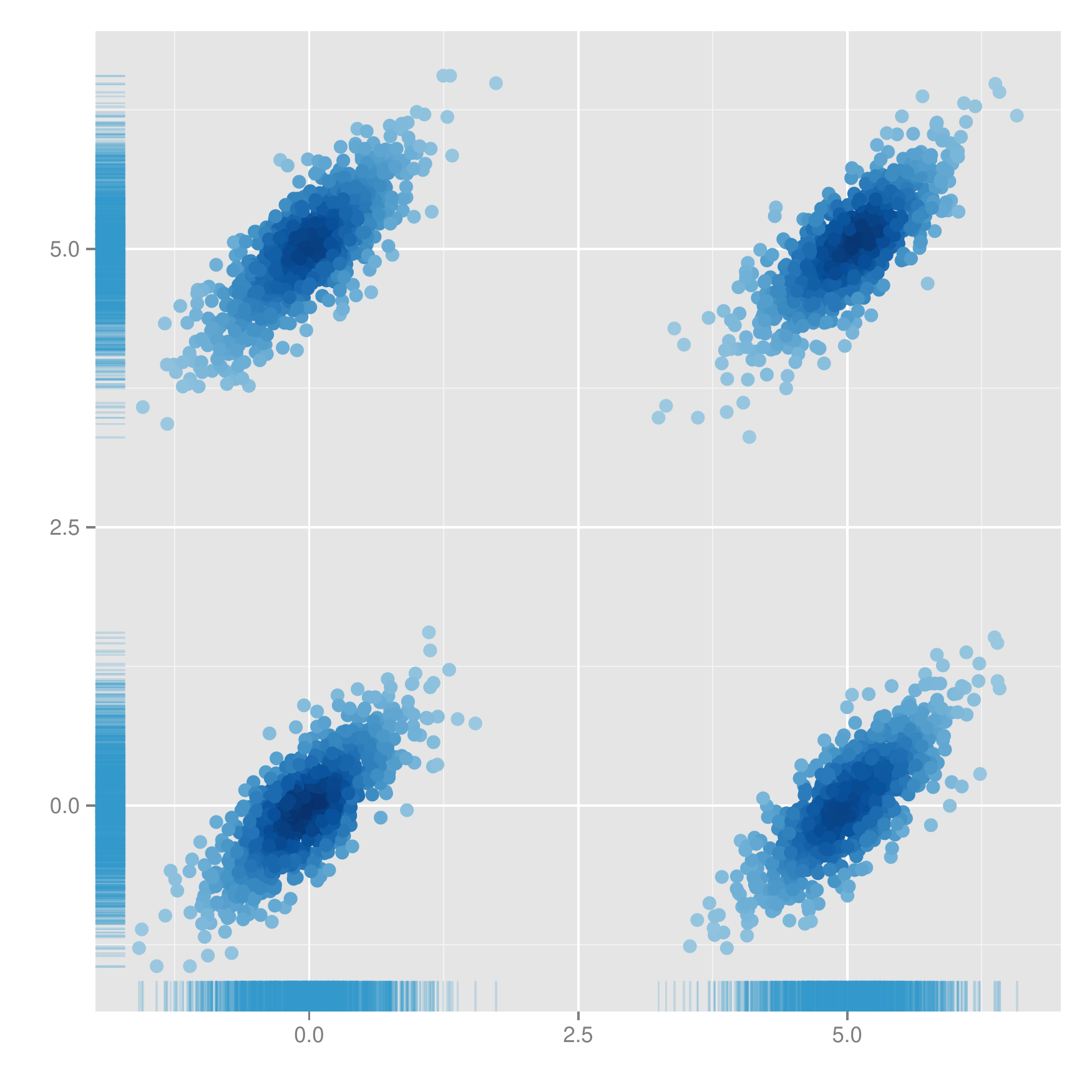}}
  \end{center}
  \caption{Left, sample from a Gaussian copula with correlation $\rho = 0.8$.
  Middle and right, two samples drawn from multivariate models with this same
  copula but different marginal distributions, depicted as rug plots.}
  \label{fig:twosamples}
\end{figure}

When the components of $\mathbf{x}=(x_1,\ldots,x_d)$ are jointly independent,
their density function $p(\mathbf{x})$ can be written as
\begin{equation}
p(\mathbf{x}) = \prod_{i=1}^{d} p(x_i)\,. \label{eq:indep}
\end{equation}
This equality does not hold when $x_1,\ldots,x_d$ are not independent.
Nevertheless, the differences can be corrected if we multiply the right hand
side of (\ref{eq:indep}) by a specific function that fully describes any
possible dependence between $x_1,\ldots,x_d$. This function is called the
\emph{copula} of $p(\mathbf{x})$ \cite{nelsen} and satisfies
\begin{equation}
p(\mathbf{x}) = \prod_{i=1}^{d} p(x_i) \, \underbrace{c(P(x_1),..., P(x_d))}_\text{copula}\,.\label{eq:copulaDensity}
\end{equation}
The copula $c$ is the joint density of $P(x_1),\ldots,P(x_d)$, where $P(x_i)$
is the marginal cdf of the random variable $x_i$.  This density has uniform
marginals, since $P(z) \sim \mathcal{U}[0,1]$ for any random variable $z$.
That is, when we apply the transformation $P(x_1),\ldots,P(x_d)$ to
$x_1,\ldots,x_d$, we are eliminating all information about the marginal
distributions.  Therefore, the copula captures any distributional pattern that
does not depend on their specific form, or, in other words, all the information
regarding the dependencies between $x_1,\ldots,x_d$.  When
$P(x_1),\ldots,P(x_d)$ are continuous, the copula $c$ is unique \cite{sklar}.
However, infinitely many multivariate models share the same underlying
copula function, as illustrated in Figure \ref{fig:twosamples}.  The main
advantage of copulas is that they allow us to model separately the marginal
distributions and the dependencies linking them together to produce the multivariate
model subject of study.

\begin{figure}
  \begin{center}
      {\includegraphics[width=0.28\linewidth]{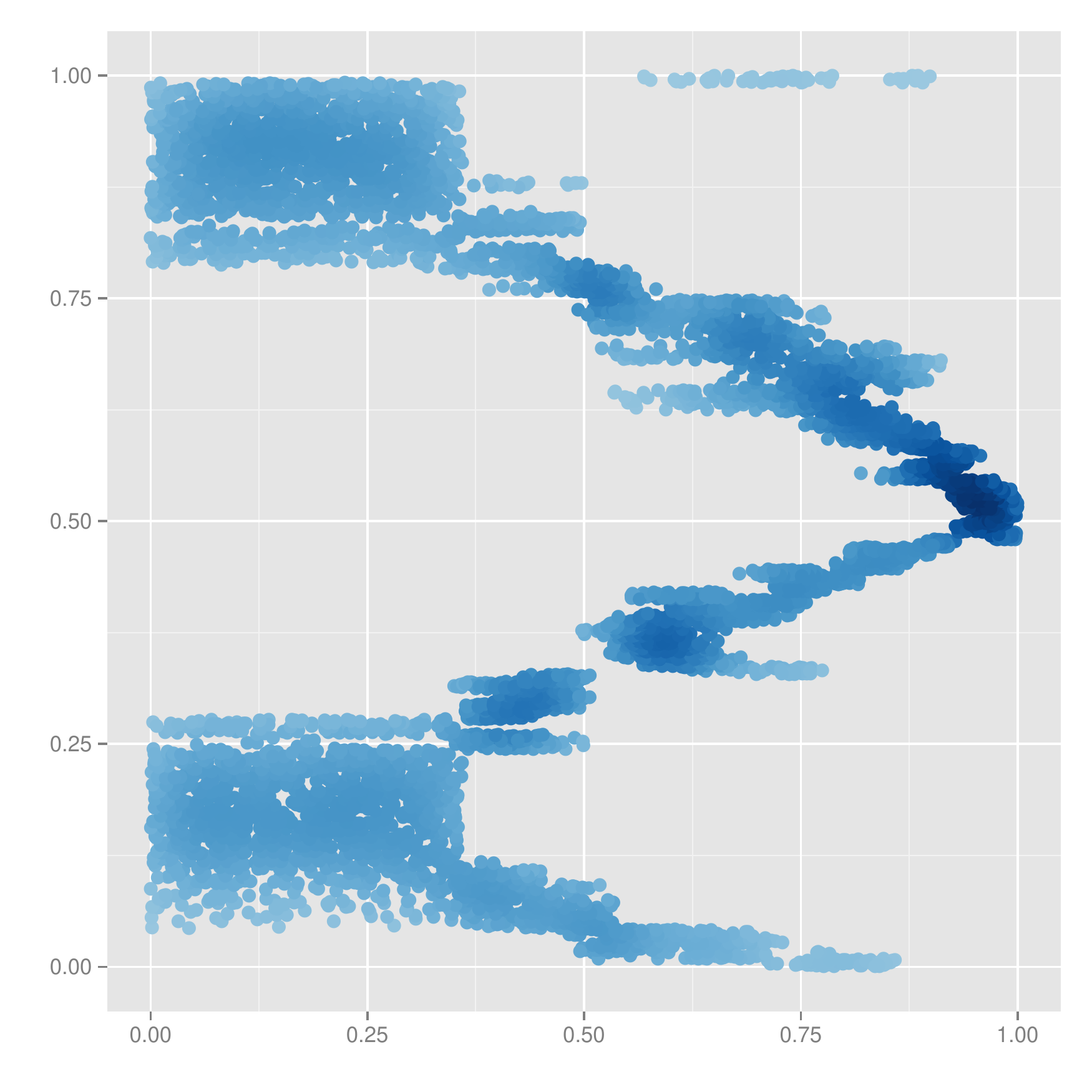}}  \hskip 1cm
      {\includegraphics[width=0.28\linewidth]{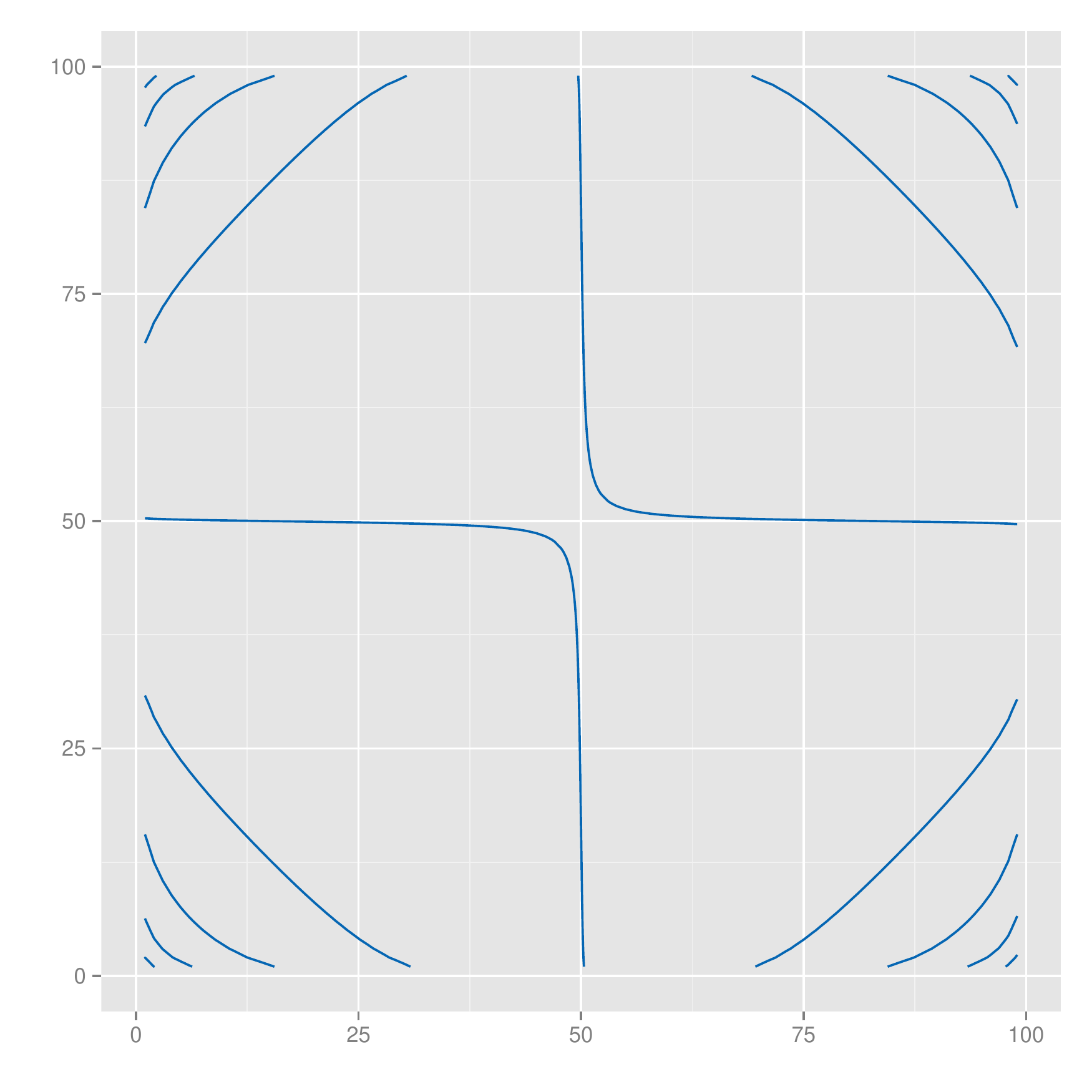}}\hskip 1cm
      {\includegraphics[width=0.28\linewidth]{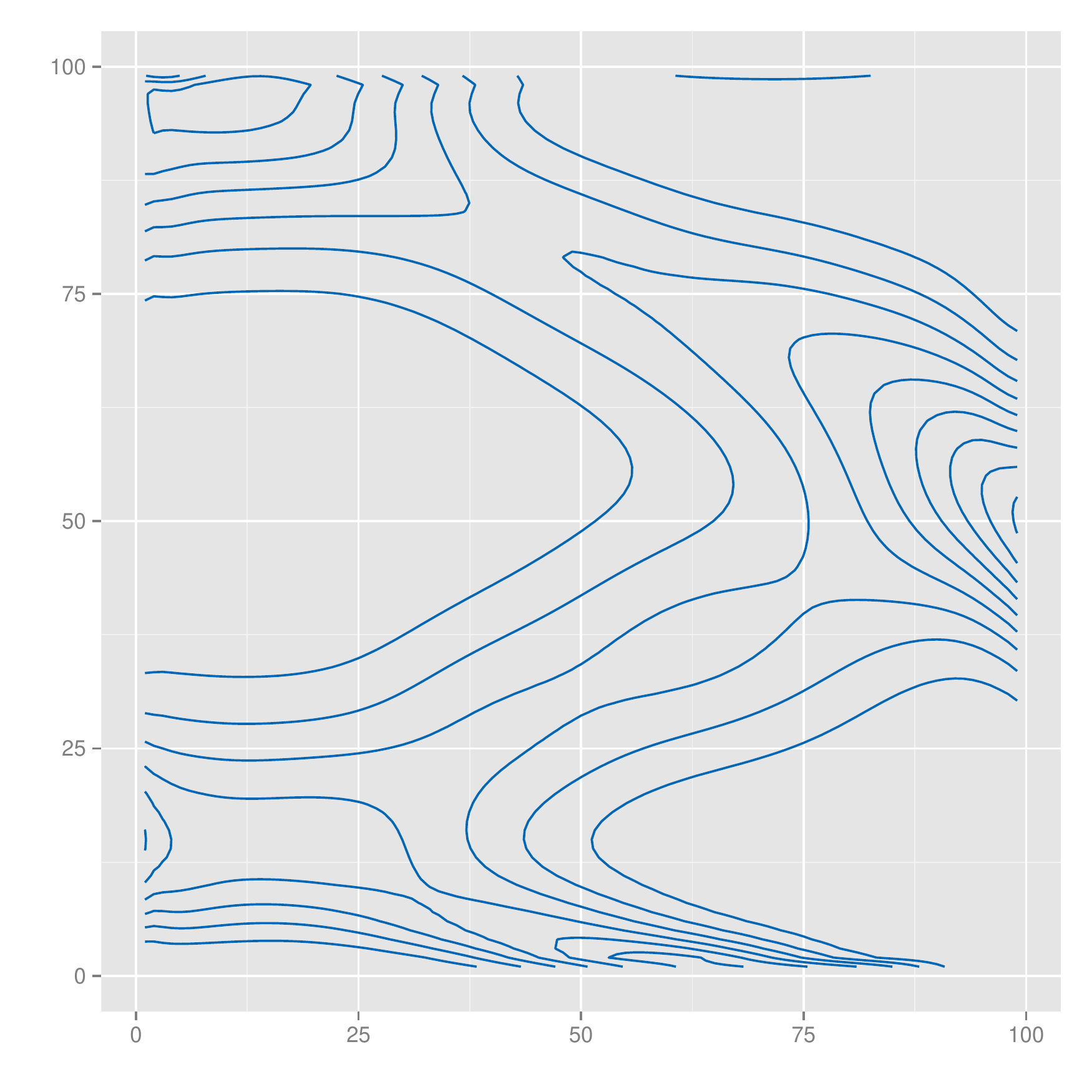}}
  \end{center}
  \caption{ Left, sample from the copula linking variables 4 and 11 in the
  \textsc{Wireless} dataset.  Middle, density estimate generated by a Gaussian
  copula model when fitted to the data.  This technique is unable to capture
  the complex patterns present in the data.  Right, copula density estimate
  generated by the non-parametric method described in section
  \ref{sec:kernel_copula}.  }
  \label{fig:complex_density}
\end{figure}

Given a sample from (\ref{eq:copulaDensity}), we can estimate $p(\mathbf{x})$
as follows. First, we construct estimates of the marginal pdfs,
$\hat{p}(x_1),\ldots,\hat{p}(x_d)$, which also provide estimates of the
corresponding marginal cdfs, $\hat{P}(x_1),\ldots,\hat{P}(x_d)$. These cdfs
estimates are used to map the data to the $d$-dimensional unit hyper-cube. The
transformed data are then used to obtain an estimate $\hat{c}$ for the copula
of $p(\mathbf{x})$.  Finally, (\ref{eq:copulaDensity}) is approximated as
\begin{equation}
\hat{p}(\mathbf{x}) = \prod_{i=1}^{d} \hat{p}(x_i) \, \hat{c}(\hat{P}(x_1),..., \hat{P}(x_d)).\label{eq:copulaApprox}
\end{equation}
The estimation of marginal pdfs and cdfs can be implemented in a non-parametric
manner by using unidimensional kernel density estimates. By contrast, it is
common practice to assume a parametric model for the estimation of the copula
function.  Some examples of parametric copulas are Gaussian, Gumbel, Frank,
Clayton or Student copulas \cite{nelsen}.  Nevertheless, real-world data often
exhibit complex dependencies which cannot be correctly described by these
parametric copula models. This lack of flexibility of parametric copulas is
illustrated in Figure \ref{fig:complex_density}.  As an alternative, we propose
to approximate the copula function in a non-parametric manner.  Kernel density
estimates can also be used to generate non-parametric approximations of
copulas, as described in \cite{kernel_copula}. The following section reviews
this method for the two-dimensional case.

\subsection{Non-parametric Bivariate Copulas}\label{sec:kernel_copula}

We now elaborate on how to non-parametrically estimate the copula of a given
bivariate density $p(x,y)$. Recall that this density can be factorized as the
product of its marginals and its copula
\begin{equation}
 p(x,y) = p(x)\,p(y)\,c(P(x),P(y))\label{eq:original}.
\end{equation}
Additionally, given a sample $\{(x_i, y_i)\}_{i=1}^n$ from $p(x,y)$, we can
obtain a pseudo-sample from its copula $c$ by mapping each observation to the
unit square using estimates of the marginal cdfs, namely
\begin{equation}
\{(u_i,v_i)\}_{i=1}^n := \{(\hat{P}(x_i), \hat{P}(y_i))\}_{i=1}^n.
\end{equation}
These are approximate observations from the uniformly distributed random
variables $u=P(x)$ and $v=P(y)$, whose joint density is the copula function
$c(u,v)$. We could try to approximate this density function by placing Gaussian
kernels on each observation $u_i$ and $v_i$.  However, the resulting density
estimate would have support on $\mathbb{R}^2$, while the support of $c$ is the
unit square.  A solution is to perform the density estimation in a transformed
space. For this, we select some continuous distribution with support on
$\mathbb{R}$, strictly positive density $\phi$, cumulative distribution $\Phi$
and quantile function $\Phi^{-1}$. Let $z$ and $w$ be two new random variables
given by $z = \Phi^{-1}(u)$ and $w = \Phi^{-1}(v)$. Then, the joint density of
$z$ and $w$ is \begin{equation} p(z,w) =
\phi(z)\,\phi(w)\,c(\Phi(z),\Phi(w))\,.\label{eq:real_density} \end{equation}
The copula of this new density is identical to the copula of
(\ref{eq:original}), since the performed transformations are marginal-wise.
The support of (\ref{eq:real_density}) is now $\mathbb{R}^2$; therefore, we can
now approximate it with Gaussian kernels.  Let $z_i = \Phi^{-1}(u_i)$ and $w_i
= \Phi^{-1}(v_i)$. Then,
\begin{equation}
\hat{p}(z,w) = \frac{1}{n} \sum_{i=1}^n \mathcal{N}(z,w| z_i, w_i, \bm \Sigma),\label{eq:approx_density}
\end{equation}
where $\mathcal{N}(\cdot,\cdot| \nu_1,\nu_2,\bm \Sigma)$ is a two-dimensional
Gaussian density with mean $(\nu_1,\nu_2)$ and covariance matrix $\bm
\Sigma$.  For convenience, we select $\phi$, $\Phi$ and $\Phi^{-1}$ to be the
standard Gaussian pdf, cdf and quantile function, respectively.  Finally, the
copula density $c(u,v)$ is approximated by combining (\ref{eq:real_density})
with (\ref{eq:approx_density}):
\begin{equation}
\hat{c}(u,v) = \frac{\hat{p}(\Phi^{-1}(u),\Phi^{-1}(v))}{\phi(\Phi^{-1}(u))\phi(\Phi^{-1}(v))} = \frac{1}{n}
\sum_{i=1}^n\frac{\mathcal{N}(\Phi^{-1}(u),\Phi^{-1}(v)| \Phi^{-1}(u_i), \Phi^{-1}(v_i), \bm \Sigma)}{\phi(\Phi^{-1}(u))\phi(\Phi^{-1}(v))}\,.
\label{eq:kernel_copula}
\end{equation}

\section{Regular Vines}\label{sec:vines}

The method described above can be generalized to the estimation of copulas of
more than two random variables.  However, although kernel density estimates can
be successful in spaces of one or two dimensions, as the number of variables
increases, this methods start to be significantly affected by the curse of
dimensionality and tend to overfit to the training data. Additionally, for
addressing domain adaptation problems, we are interested in factorizing these
high-dimensional copulas into simpler building blocks transferrable accross
learning domains.  These two drawbacks can be addressed by recent methods in
copula modelling called \emph{vines} \cite{vines}. Vines decompose any
high-dimensional copula density as a product of bivariate copula densities that
can be approximated using the non-parametric model described above.  These
bivariate copulas (as well as the marginals) correspond to the simple building
blocks that we plan to transfer from one learning domain to another.  Different
types of vines have been proposed in the literature.  Some examples are
\emph{canonical vines}, \emph{D-vines} or \emph{regular} vines
\cite{kurowicka,vines}. In this work we focus on regular vines (R-vines) since
they are the most general models.

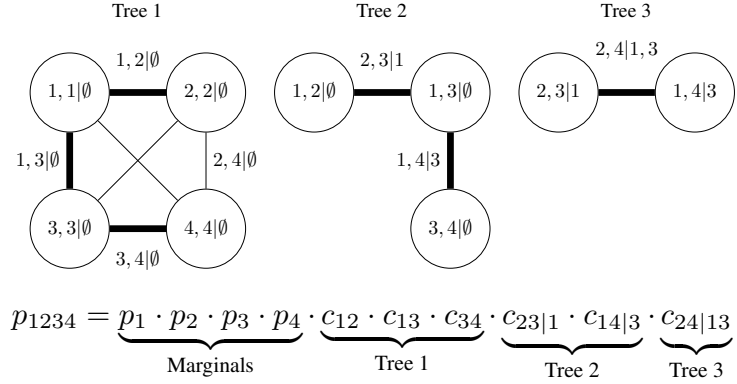
\begin{figure}[h!]
  \begin{center}
    \resizebox{0.7\linewidth}{!}
    {
      \begin{tikzpicture}
        \node[punkt]             (1) {$1,1|\emptyset$};
        \node[above=0.5 of 1,shift={(12mm,0)}] (tree1) {Tree 1};
        \node[punkt, right=of 1] (2) {$2,2|\emptyset$};
        \node[punkt, below=of 1] (3) {$3,3|\emptyset$};
        \node[punkt, below=of 2] (4) {$4,4|\emptyset$};
        \path[select] (1) edge node[anchor=south, shift={(0,2mm)}]{$1,2|\emptyset$} (2);
        \path[select] (1) edge node[anchor=east]{$1,3|\emptyset$} (3);
        \path (1) edge (4);
        \path (2) edge (3);

        \path (2) edge node[anchor=west]{$2,4|\emptyset$} (4);
        \path[select] (3) edge node[anchor=north, shift={(0,-2mm)}]{$3,4|\emptyset$} (4);
        
        \node[punkt, right=0.5 of 2] (12) {$1,2|\emptyset$};
        \node[above=0.5 of 12, shift={(12mm,0)}] (tree2) {Tree 2};

        \node[punkt, right=of 12] (13) {$1,3|\emptyset$};
        \node[punkt, below= of 13] (34) {$3,4|\emptyset$};
        \path[select] (12) edge node[anchor=south,above, shift={(0,2mm)}]{$2,3|1$} (13);
        \path[select] (13) edge node[anchor=east]{$1,4|3$} (34);

        \node[punkt, right=0.5 of 13] (231) {$2,3|1$};
        \node[above=0.5 of 231, shift={(12mm,0)}] (tree3) {Tree 3};
        \node[punkt, right=of 231,] (341) {$1,4|3$};
        \path[select] (231) edge node[anchor=south,above, shift={(0,4mm)}]{$2,4|1,3$} (341);
      \end{tikzpicture}
    }
    \begin{equation*}
      \resizebox{0.7\linewidth}{!}
      {
        $p_{1234} = \underbrace{p_1 \cdot p_2 \cdot p_3 \cdot p_4}_{\text{Marginals}}
                  \cdot \underbrace{c_{12}\cdot c_{13}\cdot c_{34}}_{\text{Tree 1}}
                  \cdot \underbrace{c_{23|1}\cdot c_{14|3}}_{\text{Tree 2}}
                  \cdot \underbrace{c_{24|13}}_{\text{Tree 3}}$
      }
      \end{equation*}
  \end{center}
\caption{Example of the hierarchical construction of a R-vine copula for a
system of four variables.  The edges selected to form each tree are highlighted
in bold. Conditioned and conditioning sets for each node and edge are shown as
$C(e) | D(e)$.  Later, each edge in bold will correspond to a different
bivariate copula function.}
\label{fig:vine_construction}
\end{figure}

An R-vine $\mathcal{V}$ for a probability density $p(x_1,\ldots,x_d)$ with
variable set $\mathbf{V} = \{1, \ldots d \}$ is formed by a set of undirected
trees $T_1, \ldots, T_{d-1}$, each of them with corresponding set of nodes
$V_i$ and set of edges $E_i$, where $V_i = E_{i-1}$ for $i \in [2, d-1]$
. Any edge $e\in E_i$ has associated three sets
$C(e),D(e),N(e)\subset\mathbf{V}$ called the conditioned, conditioning and
constraint sets of $e$, respectively. Initially, $T_1$ is inferred from a complete
graph with a node associated with each element of $\mathcal{V}$; for any $e \in T_1$ joining nodes $V_j$ and $V_k$, $C(e) = N(e) = \{V_j,V_k\} $ and $D(e) = \{ \emptyset \}$.  
The trees $T_2,...,T_{d-1}$ are constructed so that each $e
\in E_{i}$ is formed by joining two edges $e_1, e_2 \in E_{i-1}$ which share a
common node, for $i\geq2$.  The new edge $e$ has conditioned, conditioning and
constraint sets given by $C(e) = N(e_1) \Delta N(e_2)$, $D(e)= N(e_1) \cap
N(e_1)$, $N(e) = N(e_1)\cup N(e_2)$, where $\Delta$ is the symmetric
difference operator.  Figure \ref{fig:vine_construction}
illustrates this procedure for an R-vine with 4 variables.

For any edge $e(j,k)\in T_i,\,i=1,\ldots,d-1$ with conditioned set
$C(e)=\{j,k\}$ and conditioning set $D(e)$ let $c_{jk|D(e)}$ be the value of
the copula density for the conditional distribution of $x_j$ and $x_k$ when
conditioning on $\{x_i: i \in D(e) \}$, that is, 
\begin{align}
c_{jk|D(e)} & := c( P_{j|D(e)}, P_{k|D(e)}| x_i : i \in D(e)),\label{eq:defcjk}
\end{align}
where $P_{j|D(e)}:=P(x_j|x_i : i \in D(e))$ is the conditional cdf of $x_j$ when conditioning on $\{x_i: i \in D(e) \}$.
Kurowicka and Cooke \cite{kurowicka} indicate that
\emph{any} probability density function $p(x_1,\ldots,x_d)$ can then be factorized as
\begin{equation}
p(\mathbf{x}) = \prod_{i=1}^d p(x_i) \prod_{i=1}^{d-1} \prod_{e(j,k) \in E_i} c_{jk|D(e)}\,,\label{eq:vindeDecomposition}
\end{equation}
where $E_1,\ldots,E_{d-1}$ are the edge sets of the R-vine $\mathcal{V}$ for $p(x_1,\ldots,x_d)$.
In particular, each of the edges in the trees from $\mathcal{V}$ specify a different conditional
copula density in (\ref{eq:vindeDecomposition}). For $d$ variables, the density in (\ref{eq:vindeDecomposition}) is formed by
$d(d-1)/2$ factors. Changes in each of these factors 
can be detected and independently transferred accross different learning domains to improve the estimation of the target density function.

The definition of $c_{jk|D(e)}$ in (\ref{eq:defcjk}) requires the calculation of conditional marginal
cdfs. For this, we use the following recursive identity introduced by Joe \cite{joe}, that is,
\begin{equation}
P_{j|D(e)} = \frac{\partial\, C_{jk|D(e)\setminus k }}{\partial P_{k|D(e) \setminus k}}\,,\label{eq:h_function}
\end{equation}
which holds for any $k \in D(e)$, where $D(e) \setminus k = \{i: i \in D(e) \wedge i \neq k\}$ and $C_{jk|D(e)\setminus k}$ is the cdf of $c_{jk|D(e)\setminus k}$.

One major advantage of vines is that they can model high-dimensional data by
estimating density functions of only one or two random variables. For this
reason, these techniques are significantly less affected by the curse of
dimensionality than regular density estimators based on kernels, as we show in
Section \ref{sec:experiments}.  So far Vines have been generally constructed
using parametric models for the estimation of bivariate copulas.  In the
following, we describe a novel method for the construction of non-parametric
regular vines.

\subsection{Non-parametric Regular Vines}
In this section, we introduce a vine distribution in which all participant
bivariate copulas can be estimated in a non-parametric manner.  Todo so, we
model each of the copulas in (\ref{eq:vindeDecomposition}) using the
non-parametric method described in Section \ref{sec:kernel_copula}.  Let
$\{(u_i,v_i)\}_{i=1}^n$ be a sample from the copula density $c(u,v)$.  The
basic operation needed for the implementation of the proposed method is the
evaluation of the conditional cdf $P(u|v)$ using the recursive equation
(\ref{eq:h_function}).  Define $w=\Phi^{-1}(v)$, $z_i=\Phi^{-1}(u_i)$ and
$w_i=\Phi^{-1}(v_i)$.  Combining (\ref{eq:kernel_copula}) and
(\ref{eq:h_function}) we obtain
\begin{align}
\hat{P}(u|v) & = \int_{0}^u \hat{c}(x,v) \,dx \nonumber\\
&= \frac{1}{n \phi(w)} \sum_{i=1}^n{\int_{0}^{u} \frac{\mathcal{N}(\Phi^{-1}(x),w| z_i, w_i, \mathbf{\Sigma})}{\phi(\Phi^{-1}(x))}\,dx} \nonumber\\
&= \frac{1}{n\phi(w)} \sum_{i=1}^n \mathcal{N}(w| w_i, \sigma_w^2)\,\Phi\left[ \frac{\Phi^{-1}(u) - \mu_{z_i|w_i}}{\sigma^2_{z_i|w_i}} \right],\label{final_h}
\end{align}
where $\mathcal{N}(\cdot|\mu,\sigma^2)$ denotes a Gaussian density with mean
$\mu$ and variance $\sigma^2$, $\bm \Sigma = \left(\begin{array}{cc} \sigma^2_z
& \gamma\\ \gamma &\sigma^2_w\end{array}\right)$ the kernel bandwidth matrix,
$\mu_{z_i|w_i} = z_i + \frac{\sigma_z}{\sigma_w}\gamma (w - w_i)$ and
$\sigma^2_{z_i|w_i} = \sigma_z^2 (1 -\gamma^2)$.

Equation (\ref{final_h}) can be used to approximate any conditional cdf
$P_{j|D(e)}$.  For this, we use the fact that $P(x_j|x_i : i \in
D(e))=P(u_j|u_i : i \in D(e))$, where $u_i = P(x_i)$, for $i = 1,\ldots,d$, and
recursively apply rule (\ref{eq:h_function}) using equation (\ref{final_h}) to
compute $\hat{P}(u_j|u_i : i \in D(e))$.

To complete the inference recipe for the non-parametric regular vine, we must
specify how to construct the hierarchy of trees $T_1,\ldots,T_{d-1}$.  In other
words, we must define a procedure to select the edges (bivariate copulas) that
will form each tree.  We have a total of $d (d - 1) / 2$ bivariate copulas
which should be distributed among the different trees.  Ideally, we would like
to include in the first trees of the hierarchy the copulas with strongest
dependence level.  This will allow us to prune the model by assuming
independence in the last $k < d$ trees, since the density function for the
independent copula is constant and equal to 1.  To construct the trees
$T_1,\ldots,T_{d-1}$, we assign a weight to each edge $e(j,k)$ (copula)
according to the level of dependence between the random variables $x_j$ and
$x_k$. A common practice is to fix this weight to the empirical estimate of
Kendall's' $\tau$ for the two random variables under
consideration\cite{vines}\footnote{We have tried more general dependence
measures such as the HSIC (\emph{Hilbert-Schmidt Independence Criterion})
without observing gains that justify the increase of computational costs.}.
Given these weights for each edge, we propose to solve the edge selection
problem by obtaining $d-1$ maximum spanning trees. \emph{Prim's Algorithm}
\cite{prim} can be used to solve this problem efficiently.

\section{Domain Adaptation with Regular Vines}\label{sec:domain_adaptation}

In this section we describe how regular vines can be used to address domain
adaptation problems in the non-linear regression setting with continuous data.
The proposed approach could be easily extended to other problems such as
density estimation or classification.  In regression problems, we are
interested in inferring the \emph{mapping mechanism} or conditional
distribution with density $p(y|\mathbf{x})$ that maps one feature vector
$\mathbf{x}=(x_1,\ldots,x_d) \in \mathbb{R}^d$ into a target scalar value $y
\in \mathbb{R}$.  Rephrased into the copula framework, this conditional density
can be expressed as \begin{equation} p(y|\mathbf{x}) \propto p(y)
\prod_{i=1}^{d} \prod_{e(j,k) \in E_i} c_{jk|D(e)} \label{eq:regressionTarget}
\end{equation} where $E_1,\ldots,E_{d}$ are the edge sets of an R-vine for
$p(\mathbf{x},y)$.  Note that the normalization of the right part of
(\ref{eq:regressionTarget}) is relatively easy since $y$ is scalar.

In the classic domain adaptation setup we usually have large amounts of data
for solving a source task characterized by the density function
$p_s(\mathbf{x}, y)$. However, only a partial or reduced sample is available
for solving a target task with density $p_t(\mathbf{x}, y)$. Given the data
available for both tasks, our objective is to build a good estimate for the
conditional density $p_t(y | \mathbf{x})$.  To address this domain
adaptation problem, we assume that $p_t$ is a modified version of $p_s$.
In particular, we assume that $p_t$ is obtained in two steps from $p_s$.
First, $p_s$ is expressed using an R-vine representation as in
(\ref{eq:vindeDecomposition}) and second, some of the factors included in
that representation (marginal distributions or pairwise copulas) are
modified to derive $p_t$.  All we need to address the adaptation across
domains is to reconstruct the R-vine representation of $p_s$ using data
from the source task, and then identify which of the factors have been
modified to produce $p_t$. These factors are corrected using data from the
target task. In the following, we describe how to identify and correct
these modified factors.

Marginal distributions can change between source and target tasks (also known
as \emph{covariate shift}).  In this case, $P_s(x_i) \neq P_t(x_i)$, for
$i=1,\ldots,d$, or $P_s(y) \neq P_t(y)$, and we need to re-generate the
estimates of the affected marginals using data from the target task.
Additionally, some of the bivariate copulas $c_{jk|D(e)}$ may differ from
source to target tasks.  In this case, we also re-estimate the affected copulas
using data from the target task.  Simultaneous changes in both copulas and
marginals can occur. However, there is no limitation in updating each of the
modified components separately.  Finally, if some of the factors remain
constant across domains, we can use the available data from the target task to
improve the estimates obtained using only the data from the source task.  Note
that we are addressing a more general problem than \emph{covariate shift}.
Besides identifying and correcting changes in marginal distributions, we also
consider changes in any possible form of dependence (conditional distributions)
between random variables.

For the implementation of the strategy mentioned above, we need to identify
when two samples come from the same distribution or not. For this, we propose
to use the non-parametric two-sample test \emph{Maximum Mean Discrepancy} (MMD)
\cite{mmd}.  MMD will return low $p$-values when two samples are unlikely to
have been drawn from the same distribution.  Specifically, given samples from
two distributions $P$ and $Q$, MMD will determine $P\neq Q$ if the distance
between the embeddings of the empirical distributions for these two samples in
a RKHS is significantly large.

\paragraph{Semi-supervised and unsupervised domain adaptation:}
The proposed approach can be easily extended to take advantage of additional
unlabeled data to improve the estimation of our model. Specifically, extra
unlabeled target task data can be used to refine the factors in the R-Vine
decomposition of $p_t$ which do not depend on $y$.  This is still valid even in
the limiting case of not having access to labeled data from the target task at
training time (\emph{unsupervised domain adaptation}).

\section{Experiments}\label{sec:experiments}
To validate the proposed method, we run two series of experiments using real
world data.  The first series illustrates the accuracy of the density estimates
generated by the proposed non-parametric vine method.  The second series
validates the effectiveness of the proposed framework for domain adaptation
problems in the non-linear regression setting.  In all experiments, kernel
bandwidth matrices are selected using Silverman's rule-of-thumb
\cite{silverman}.  For comparative purposes, we include the results of
different state-of-the-art domain adaptation methods whose parameters are
selected by a 10-fold cross validation process on the training data.

\paragraph{Approximations:} A complete R-Vine requires the use of conditional
copula functions, which are challenging to learn.  A common approximation is to
ignore any dependence between the copula functional form and its set of
conditioning variables. Note that the copula functions arguments remain to be
conditioned cdfs. Moreover, to avoid excesive computational costs, we consider
only the first tree ($d-1$ copulas) of the R-Vine, which is the one containing
the most amount of dependence between the distribution variables. Increasing
the number of considered trees did not lead to significant performance
improvements.

\subsection{Accuracy of Non-parametric Regular Vines for Density Estimation}
\label{sec:exp_vines}

The density estimates generated by the new non-parametric R-vine method (NPRV)
are evaluated on data from six normalized UCI datasets \cite{uci}.  We compare
against a standard density estimator based on Gaussian kernels (\textsc{KDE}), 
and a parametric vine method based on bivariate Gaussian copulas
(\textsc{GRV}).  From each dataset, we extract 50 random samples of size 1000.
Training is performed using 30\% of each random sample.  Average test
log-likelihoods and corresponding standard deviations on the remaining 70\% of
the random sample are summarized in Table \ref{table:vine_exps} for each
technique.  In these experiments, \textsc{NPRV} obtains the highest average
test log-likelihood in all cases except one, where it is outperformed by
\textsc{GRV}. \textsc{KDE} shows the worst performance, due to its
direct exposure to the curse of dimensionality.

\begin{table}
\begin{center}
\caption{Average TLL obtained by \textsc{NPRV}, \textsc{GRV} and \textsc{KDE} on six different UCI datasets.}
\resizebox{\linewidth}{!}
{
\begin{tabular}{lcccccc}
\hline
Dataset & \textbf{Auto} & \textbf{Cloud} & \textbf{Housing} & \textbf{Magic} & \textbf{Page-Blocks} & \textbf{Wireless}\\
\hline
No. of variables & \textbf{8} & \textbf{10} & \textbf{14} & \textbf{11} & \textbf{10} & \textbf{11}\\ \hline
\textbf{KDE} & 1.32 $\pm$ 0.06 & 3.25 $\pm$ 0.10& 1.96 $\pm$ 0.17 & 1.13 $\pm$ 0.11& 1.90 $\pm$ 0.13& 0.98 $\pm$ 0.06\\ 
\textbf{GRV}  & 1.84 $\pm$ 0.08 & \bf{5.00} $\pm$ \bf{0.12} & 1.68 $\pm$ 0.11 & 2.09 $\pm$ 0.08& 4.69 $\pm$ 0.20& 0.36 $\pm$ 0.08\\ 
\textbf{NPRV} & \bf{2.07} $\pm$ \bf{0.07} & 4.54 $\pm$ 0.13& \bf{3.18} $\pm$ \bf{0.17} & \bf{2.72} $\pm$ \bf{0.17} & \bf{5.64} $\pm$ 0.14& \bf{2.17} $\pm$ 0.13\\ 
\hline
\label{table:vine_exps}
\end{tabular}
}
\end{center}
\end{table}

\subsection{Comparison with other Domain Adaptation Methods}

\textsc{NPRV} is analyzed in a series of experiments for domain adaptation on
the non-linear regression setting with real-world data.  Detailed descriptions
of the 6 UCI selected datasets and their domains are available in the
supplementary material.  The proposed technique is compared with different
benchmark methods.  The first two, \textsc{GP-Source} and \textsc{GP-All}, are
considered baselines.  They are two gaussian process (GP) methods, the first
one trained only with data from the source task, and the second one trained
with the normalized union of data from both source and target problems.  The
other five methods are considered state-of-the-art domain adaptation
techniques.  \textsc{Daume} \cite{daume} performs a feature augmentation such
that the kernel function evaluated at two points from the same domain is twice
larger than when these two points come from different domains.
\textsc{SSL-Daume}  \cite{ssldaume} is a SSL extension of \textsc{Daume} which
takes into account unlabeled data from the target domain.  \textsc{ATGP}
\cite{bin} models the source and target task data using a single GP, but learns
additional kernel parameters to correlate input vectors between domains. This
method outperforms others like the one proposed by Bonilla et al.
\cite{bonilla}.  \textsc{KMM} \cite{kmm} minimizes the distance of marginal
distributions in source and target domains by matching their means when mapped
into an universal RKHS. Finally, \textsc{KuLSIF} \cite{kulsif} operates in a
similar way as \textsc{KMM}.  Besides \textsc{NPRV}, we also include in the
experiments its fully unsupervised variant, \textsc{UNPRV}, which ignores any
labeled data from the target task.

For training, we randomly sample 1000 data points for both source and target
tasks, where all the data in the source task and 5\% of the data in the target
task are labeled.  The test set contains 1000 points from the target task.
Table \ref{table:da_exps} summarizes the average test normalized mean square
error (NMSE) and corresponding standard deviation for each method in each
dataset across 30 random repetitions of the experiment.  The proposed methods
obtain the best results in 5 out of 6 cases. Notably, \textsc{UNPRV}
(Unsupervised \textsc{NPRV}), which ignores labeled data from the target task,
also outperforms the other benchmark methods in most cases.  Finally, the two
bottom rows in Table \ref{table:da_exps} show the average number of marginals
and bivariate copulas which are updated in each dataset during the execution of
\textsc{NPRV}, respectively.
\begin{table}
  \begin{center}
  \caption{Average NMSE and standard deviation for all algorithms and UCI datasets.}
  \resizebox{\linewidth}{!}
  {
  \begin{tabular}{lcccccc}
  \hline
    Dataset & \textbf{Wine}& \textbf{Sarcos}& \textbf{Rocks-Mines}& \textbf{Hill-Valleys}&\textbf{Axis-Slice}&\textbf{Isolet}  \\
  \hline
    No. of variables & \textbf{12}& \textbf{21}& \textbf{60}& \textbf{100}&\textbf{386}&\textbf{617}  \\ \hline
 %                        Wine*               Sarcos*                Sonar                Hills              CT-Slice            Isolet
 \f{GP-Source} &    0.86 $\pm$ 0.02  &    1.80 $\pm$ 0.04  &    0.90 $\pm$ 0.01  &    1.00 $\pm$ 0.00  &    1.52 $\pm$ 0.02  &    1.59 $\pm$ 0.02 \\ 
 \f{GP-All   } &    0.83 $\pm$ 0.03  &    1.69 $\pm$ 0.04  &    1.10 $\pm$ 0.08  &    0.87 $\pm$ 0.06  &    1.27 $\pm$ 0.07  &    1.58 $\pm$ 0.02 \\ 
 \f{Daume    } &    0.97 $\pm$ 0.03  &    0.88 $\pm$ 0.02  &    0.72 $\pm$ 0.09  &    0.99 $\pm$ 0.03  &    0.95 $\pm$ 0.02  &    0.99 $\pm$ 0.00 \\ 
 \f{SSL-Daume} &    0.82 $\pm$ 0.05  &    0.74 $\pm$ 0.08  &    0.59 $\pm$ 0.07  &    0.82 $\pm$ 0.07  &    0.65 $\pm$ 0.04  &    0.64 $\pm$ 0.02 \\ 
 \f{ATGP     } &    0.86 $\pm$ 0.08  &    0.79 $\pm$ 0.07  & \f{0.56 $\pm$ 0.10} &    0.15 $\pm$ 0.07  &    1.00 $\pm$ 0.01  &    1.00 $\pm$ 0.00 \\ 
 \f{KMM      } &    1.03 $\pm$ 0.01  &    1.00 $\pm$ 0.00  &    1.00 $\pm$ 0.00  &    1.00 $\pm$ 0.00  &    1.00 $\pm$ 0.00  &    1.00 $\pm$ 0.00 \\ 
 \f{KuLSIF   } &    0.91 $\pm$ 0.08  &    1.67 $\pm$ 0.06  &    0.65 $\pm$ 0.10  &    0.80 $\pm$ 0.11  &    0.98 $\pm$ 0.07  &    0.58 $\pm$ 0.02 \\ 
 \f{NPRV     } & \f{0.73 $\pm$ 0.07} & \f{0.61 $\pm$ 0.10} &    0.72 $\pm$ 0.13  & \f{0.15 $\pm$ 0.07} &   {0.38 $\pm$ 0.07} &    0.46 $\pm$ 0.09 \\ 
 \f{UNPRV    } &    0.76 $\pm$ 0.06  &    0.62 $\pm$ 0.13  &    0.72 $\pm$ 0.15  &    0.19 $\pm$ 0.09  & \f{0.37 $\pm$ 0.07} & \f{0.42 $\pm$ 0.04} \\
    \hline
  Av. Ch. Mar. & 10 & 1 & 38 & 100 & 226 & 89\\
  Av. Ch. Cop. & 5  & 8 & 49 & 34 & 155 & 474\\
  \hline
  \label{table:da_exps}
  \end{tabular}
  }
\end{center}
\end{table}

\paragraph{Computational Costs:} Running \textsc{NPRV} requires to fill in a
weight matrix of size $\mathcal{O}(d^2)$ with the empirical estimates of
Kendall's $\tau$ for any two random variables.  The computation of each of
these estimates can be done efficiently with cost $\mathcal{O}(n\log n)$, where
$n$ is the number of available data points. Therefore, the final training cost
of \textsc{NPRV} is $\mathcal{O}(d^2 n \log n)$.  In practice, we obtain
competitive training times. Training \textsc{NPRV} for the  \emph{Isolet}
dataset took about 3 minutes on a regular laptop computer. Predictions made by
a single level \textsc{NPRV} have cost $\mathcal{O}(nd)$. Parametric copulas
may be used to reduce the computational demands.

\section{Conclusions}
We have proposed a novel non-parametric domain adaptation strategy based on
copulas.  The new approach works by decomposing any multivariate density into a
product of marginal densities and bivariate copula functions. Changes in these
factors across different domains can be detected using two sample tests, and
transferred across domains in order to adapt the target task density model.  A
novel non-parametric vine method has been introduced for the practical
implementation of this method. This technique leads to better density estimates
than standard parametric vines or KDE, and is also able to outperform a large
number of alternative domain adaptation methods in a collection of regression
problems with real-world data.

%\paragraph{Future work directions:} extend our method to classification
%problems, implement mechanisms of soft substitution based on the returned
%statistics by the two-sample tests.

\newpage

\bibliography{Bibliography}
\bibliographystyle{plain}
\end{document}